# Methods for Computing State Similarity in Markov Decision Processes

**Norm Ferns, Pablo Samuel Castro, Doina Precup, Prakash Panangaden**
School of Computer Science
McGill University
Montréal, Canada, H3A 2A7
{nferns, pcastr, dprecup, prakash}@cs.mcgill.ca

## Abstract

A popular approach to solving large probabilistic systems relies on aggregating states based on a measure of similarity. Many approaches in the literature are heuristic. A number of recent methods rely instead on metrics based on the notion of bisimulation, or behavioral equivalence between states (Givan et al., 2003; Ferns et al., 2004). An integral component of such metrics is the Kantorovich metric between probability distributions. However, while this metric enables many satisfying theoretical properties, it is costly to compute in practice. In this paper, we use techniques from network optimization and statistical sampling to overcome this problem. We obtain in this manner a variety of distance functions for MDP state aggregation that differ in the trade-off between time and space complexity, as well as the quality of the aggregation. We provide an empirical evaluation of these tradeoffs.

## 1 Introduction

Markov decision processes (MDPs) are the model of choice for decision making under uncertainty (Boutilier et al., 1999), and provide a standard formalism for describing multi-stage decision making in probabilistic environments. The objective of the decision making is to maximize a cumulative measure of long-term performance, called the *return*. Dynamic programming algorithms, e.g., value iteration or policy iteration (Puterman, 1994), allow computing the optimal expected return for any state, as well as the way of behaving (policy) that generates this return. However, in many practical situations, the state space of an MDP is too large for the standard algorithms to be applied. One popular technique for overcoming this problem is state aggregation: states are grouped together into blocks, or partitions, and a new MDP is defined over these. The hope is that this can be done in such a manner as to construct an "essentially equivalent" MDP with drastically reduced state space, thereby allowing the use of classical solution methods, while at the same time providing a guarantee that solutions of the reduced MDP can be extended to the original model.

It has been well argued that the notion of "essentially equivalent" in probabilistic systems is perhaps best captured formally by bisimulation (Milner, 1980; Park, 1981; Larsen & Skou, 1991). In the context of MDPs, bisimulation can roughly be described as the largest equivalence relation on the state space of an MDP that relates two states precisely when for every action, they achieve the same immediate reward and have the same probability of transitioning to classes of equivalent states. This means that bisimilar states lead to essentially the same long-term behavior. The bisimulation equivalence classes can even be computed iteratively in polynomial time (Givan et al., 2003). However, it has also been well established that using exact equivalences in probabilistic systems is problematic. A notion of equivalence is two-valued: two states are either equivalent or not equivalent. A small perturbation of the transition probabilities can make two equivalent states no longer equivalent. In short, any kind of equivalence is too unstable, too sensitive to perturbations of the numerical values of the transition probabilities.

A natural remedy is to use (pseudo)metrics. Metrics are natural quantitative analogues of the notion of equivalence relation. For example, the triangle inequality is a natural quantitative analogue of transitivity. The metrics on which we focus specify the degree to which objects of interest behave similarly. In Ferns et al. (2004; 2005), based on similar work in the context of labeled Markov processes (Desharnais et al., 1999; van Breugel and Worrell 2001a; 2001b), we sought to extend bisimulation for MDPs quantitatively in terms of such metrics. Our metrics, based on the Kantorovich probability metric, indeed not only provide the appropriate generalization of bisimulation, but satisfy many nice additional theoretical properties as well. Unfortunately, this integral component, the Kantorovich metric, is also one that makes our metrics expensive to compute in practice.

In this paper, we explore ways of obtaining practical distance metrics through efficient computation and approximation of the Kantorovich metric. We use techniques from optimization and sampling to obtain variations on our bisimulation metrics that are easier to compute in practice while still maintaining theoretical guarantees.

The paper is organized as follows. In Section 2 we provide the notation and theoretical background necessary for understanding the problem at hand. In Section 3 we introduce various candidate state similarity metrics and discuss the merits and drawbacks of each. We provide experiments in Section 4, to compare and contrast these. Finally, Section 5 contains conclusions and directions for future work.

## 2 Background

### 2.1 Markov Decision Processes

A finite Markov decision process is a quadruple $(S, A, \{P_{ss'}^a\}, \{r_s^a\})$ where $S$ is a finite set of states, $A$ is a finite set of actions, $P_{ss'}^a$ is a (Markovian) probability of transitioning from state $s$ to $s'$ under action $a$, and $r_s^a$ is the numerical reward for choosing action $a$ in state $s$.

The discounted, infinite horizon planning task in an MDP is to determine a policy $\pi : S \to A$ that maximizes the value of every state, $V^\pi(s) = \mathbb{E}[\sum_{t=0}^\infty \gamma^t r_{t+1} | s_0 = s, \pi]$, where $s_0$ is the state at time 0, $r_{t+1}$ is the reward achieved at time $t+1$, $\gamma$ is a discount factor in $(0,1)$, and the expectation is taken by following the state dynamics induced by $\pi$. The function $V^\pi$ is called the value function of policy $\pi$. The optimal value function $V^*$, associated with an optimal policy, is the unique solution of the set of equations:

$$V^*(s) = \max_{a \in A}(r_s^a + \gamma \sum_{s' \in S} P_{ss'}^a V^*(s'))$$

and can be used to determine an optimal policy, by choosing actions greedily. In fact, the optimal value function can be computed as the limit of a sequence of iterates (Puterman, 1994, theorem 6.2.12) . Define $V_0 = 0$ and $V_{n+1}(s) = \max_{a \in A}(r_s^a + \gamma \sum_{s' \in S} P_{ss'}^a V_n(s'))$. Then $\{V_n\}$ converges uniformly to $V^*$.

These results can be realized via a dynamic programming (DP) algorithm; however, it is often the case that the state space of the given MDP is too large for DP to be feasible. When this happens, a standard strategy is to approximate the given MDP by aggregating its state space. The hope is that one can obtain a smaller "equivalent" MDP with an easily computable value function that could provide information about the value function of the original MDP. Givan et al. (2003) investigated several notions of MDP state equivalence and determined that the most appropriate is bisimulation.

**Definition 2.1.** Bisimulation is the largest equivalence relation $\sim$ on $S$ satisfying the following property:

$$s \sim s' \iff \forall a \in A, (r_s^a = r_{s'}^a \text{ and}$$
$$\forall C \in S/\sim, P_s^a(C) = P_{s'}^a(C))$$

Unfortunately, as an exact equivalence, bisimulation suffers from issues of instability; that is, slight numerical differences in the MDP parameters, $\{r_s^a\}$ and $\{P_{ss'}^a\}$, can lead to vastly different bisimulation partitions. To get around this, one generalizes the notion of bisimulation equivalence through bisimulation metrics.

### 2.2 Bisimulation Metrics

In Ferns et al. (2004; 2005) we provide the following metric generalization of bisimulation [1]:

**Theorem 2.2.** *Let $c \in (0,1)$ and $\mathcal{V}$ be the set of real-valued functions on $S \times S$ that are bounded by $\frac{R}{(1-c)}$, where $R = \max_{s,s',a} |r_s^a - r_{s'}^a|$. Define $F^c : \mathcal{V} \to \mathcal{V}$ by*

$$F^c(h)(s,s') = \max_{a \in A}(|r_s^a - r_{s'}^a| + cT_K(h)(P_s^a, P_{s'}^a))$$

*Then :*

1. *$F^c$ has a unique fixed point $d_{fix}^c$,*

2. *$d_{fix}^c(s, s') = 0 \iff s \sim s'$, and*

3. *for any $h_0 \in \mathcal{V}$,*

$$\|d_{fix}^c - (F^c)^n(h_0)\| \leq \frac{c^n}{1-c}\|F^c(h_0) - h_0\|.$$

Here $T_K(h)(P, Q)$ is the Kantorovich probability metric[2] applied to distributions $P$ and $Q$. It is defined as $\min_\lambda \mathbb{E}_\lambda[h]$ where $\lambda$ is a joint probability function on $S \times S$ with projections $P$ and $Q$, i.e.

$$\min_{\lambda_{kj}} \sum_{k,j=1}^{|S|} \lambda_{kj} h(s_k, s_j)$$
$$\text{subject to: } \forall k. \sum_j \lambda_{kj} = P(s_k)$$
$$\forall j. \sum_k \lambda_{kj} = Q(s_j)$$
$$\forall k, j. \lambda_{kj} \geq 0$$

This formulation shows that this metric is an instance of the minimum cost flow (MCF) network optimization linear program (LP). Since there exist strongly polynomial algorithms to compute the MCF problem (Orlin, 1988), the

---
[1]Results appear here in slightly modified form.
[2]Frustratingly also known as Monge-Kantorovich, Kantorovich-Rubinstein, Hutchinson, Mallows, Wasserstein, Vasserstein, Earth Mover's Distance, and more!

Kantorovich metric can be computed in polynomial time. For our purposes, this amounts to a worst case running time of $O(|S|^3 \log |S|)$ for each Kantorovich LP (contrast this with the general LP for directly computing the optimal value function: this has $|S|$ variables and $|A||S|$ constraints (Puterman, 1994)).

The key property of the Kantorovich metric is that it matches distributions, i.e. assigns them distance zero only when they agree on the equivalence classes induced by the underlying cost function. Therefore, it is not surprising that it can be used to capture the notion of bisimulation, which requires that probabilistic transitions agree on bisimulation equivalence classes.

The Kantorovich metric also admits a characterization in terms of the coupling of random variables. We may write $T_K(h)(P,Q) = \min_{(X,Y)} \mathbb{E}[h(X,Y)]$ where the expectation is taken with respect to the joint distribution of $(X,Y)$ and the minimum is taken with respect to all pairs of random variables $(X,Y)$ such that the marginal distribution of $X$ is $P$ and the marginal distribution of $Y$ is $Q$.

## 3 State Similarity Metrics

### 3.1 Fixed Point

The metric defined by theorem 2.2 can be re-written as:
$$d_{fix}^c(s,s') = \max_{a \in A}(|r_s^a - r_{s'}^a| + cT_K(d_{fix}^c)(P_s^a, P_{s'}^a)).$$

Each such metric is continuous in the MDP parameters $\{r_s^a\}$ and $\{P_{ss'}^a\}$ and admits tight bounds on the optimal value function, since the optimal value function with discount factor $\gamma$ is Lipschitz-continuous with respect to each metric satisfying $c \geq \gamma$. Thus, when using this metric to aggregate states, it is easy to address issues of instability and being able to recover optimal solutions. Moreover, the theorem provides a way of calculating $d_{fix}^c$: starting with the metric that is zero everywhere, iteratively apply $F^c$ until a prescribed degree of accuracy is achieved. Unfortunately, directly computing the Kantorovich metric at each iteration is too costly in practice, severely limiting the use of the fixed point metrics.

### 3.2 Fixed Point with Cost Reoptimization

One way of overcoming the costly computation of the Kantorovich LP for every iteration is to use cost reoptimization. The idea is that if $h_0$ is close to $h_1$ in (uniform norm) distance then optimal solutions to $T_K(h_0)(P,Q)$ and $T_K(h_1)(P,Q)$ should be close too; so instead of starting a network optimization algorithm for $T_K(h_1)(P,Q)$ from scratch we save the optimal solution to $T_K(h_0)(P,Q)$ and use it as the starting solution[3]. In a sense, we are comput-ing one MCF LP with cost function $d_{fix}^c$. This idea has been rather extensively and successfully explored in (Frangioni & Manca., 2006). Note that any savings in time comes at the cost of space requirements, as we are now required to save solutions for each Kantorovich LP between iterations.

### 3.3 Fixed Point with Sampling

A more promising approach is a quick and efficient approximation arising from statistical sampling. Suppose $P$ and $Q$ are approximated using the empirical distributions $P_i$ and $Q_i$. That is, we sample $i$ points $X_1, X_2, \ldots, X_i$ independently according to $P$ and define $P_i$ by $P_i(x) = \frac{1}{i} \sum_{k=1}^{i} \delta_{X_k}(x)$. Similarly, write $Q_i(x) = \frac{1}{i} \sum_{k=1}^{i} \delta_{Y_k}(x)$. Then

$$T_K(h)(P_i, Q_i) = \min_{\sigma} \frac{1}{i} \sum_{k=1}^{i} h(X_k, Y_{\sigma(k)}) \quad (1)$$

where the minimum is taken over all permutations $\sigma$ on $i$ elements (see p. 12 of Villani (2002); this is a consequence of Birkhoff's theorem). Now the Strong Law of Large Numbers (SLLN) tells us that both $\{P_i(x)\}$ and $\{Q_i(x)\}$ converge almost surely to $P(x)$ and $Q(x)$[4]. Let us write $T_K^i(h)(P,Q)$ for $T_K(h)(P_i, Q_i)$ when the empirical distributions are fixed. Then as a consequence of the SLLN, $\{T_K^i(h)(P,Q)\}$ converges to $T_K(h)(P,Q)$ almost surely; moreover replacing $T_K$ by $T_K^i$ in $F^c$ yields a metric,

$$d_i^c(s,s') = \max_{a \in A}(|r_s^a - r_{s'}^a| + cT_K^i(d_i^c)(P_s^a, P_{s'}^a)),$$

which converges to $d_{fix}^c$ as $i$ gets large (see appendix A for an outline of the proof).

The importance of this result stems from the fact that the expression in equation (1) is an instance of the assignment problem from network optimization. This is a special form of the MCF problem in which the underlying network is bipartite and all flow assignments are either 0 or 1[5]. Its specialized structure allows for simpler, faster solution methods. For example, the Hungarian algorithm (for a description see Munkres (1957)) runs in worst case time $O(i^3)$, where $i$ is the number of samples.

For the continuous space $\mathbb{R}$ with the usual Euclidean metric this approximation of the Kantorovich distance is commonly known as the empirical Mallows distance. It is used to test equivalence of empirical distributions in statistics.

### 3.4 Total Variation

The standard metric for measuring the distance between probability functions is the total variation metric, defined by $TV(P,Q) = \frac{1}{2} \sum_{s \in S} |P(s) - Q(s)|$, which is half the $L^1$- norm of $P-Q$. It is a strong measure of convergence, in the sense that distributions will have distance zero only when they agree exactly on transitions to *every state*. In contrast, the Kantorovich metric demands agreement only on *classes* of states. Nevertheless, the total variation is a simple concept and one that is easy to compute.

---

[3]In LP jargon this concept is known as sensitivity analysis. Convergence of the iterates $(F^c)^n(h_0)$ to $d_{fix}^c$ in uniform norm guarantees that it applies here.

[4]Note that both $P_i$ and $Q_i$ are random variables.

[5]In graph theoretic terminology, this is the problem of optimal matching in a weighted graph.

In Ferns et al. (2004), we suggested that in place of iteratively applying $F^c$ to an initial metric $h_0$ until convergence to $d^c_{fix}$, we start with an appropriately chosen $h_0$ and apply $F^c$ only once. If we take $h_0$ to be equal to the discrete metric assigning distance $\frac{R}{(1-c)}$[6] to all pairs of unequal states then the resultant metric is

$$d^c_{TV}(s,s') = \max_{a \in A}(|r^a_s - r^a_{s'}| + \frac{cR}{1-c}TV(P^a_s, P^a_{s'})).$$

This is in some sense the simplest metric one can compute, yet the one also providing the least guarantees (at the other extreme lies $d^c_{fix}$). On the other hand, if we take $h_0$ to be equal to the discrete metric assigning distance $\frac{R}{(1-c)}$ to all pairs of states that are not bisimilar, then the resulting metric is

$$d^c_\sim(s,s') = \max_{a \in A}(|r^a_s - r^a_{s'}| + \frac{cR}{1-c}TV_\sim(P^a_s, P^a_{s'})),$$

whose probability metric component $TV_\sim(P,Q) := \frac{1}{2}\sum_{C \in S/\sim}|P(C) - Q(C)|$ is the total variation distance defined with respect to the minimized bisimilar MDP. It is relatively simply to compute, requiring only the computation of the exact bisimulation partition. However, as this latter component is unstable, so is the resultant metric. More precisely, this metric is not continuous in the MDP parameters.

Note that by monotonicity of the functional $T_K$, we immediately have $d^c_{fix} \leq d^c_\sim \leq d^c_{TV}$.

## 4 Experiments

Experiments were run on four different MDPs: a $3 \times 3$ grid world with two actions (move forward and rotate) and a single reward in the center of the room; a $5 \times 5$ and a $7 \times 7$ grid world each with the same dynamics; and a flattened out version of the coffee robot MDP (Boutilier et al., 1995) where the robot has to get coffee for the user and avoid getting wet. Note that we chose on purpose small environments, which would allow a thorough study of all the properties of the metrics. For all gridworlds, the state includes both the position as well as the orientation of the agent. So the three gridworlds have 36, 100 and 196 states respectively. The actions are deterministic. The coffee domain has 64 states and 4 actions, some with stochastic effects. Five methods were used to compute distances for these MDPs: $d^c_{fix}$, $d^c_{fix}$ with cost reoptimization, $d^c_i$ via sampling, $d^c_{TV}$, and $d^c_\sim$.

[6]This scaling factor is added to ensure that $h_0$ belongs to $\mathcal{V}$.

The first two methods find the fixed point metric by computing the distance between two distributions through the Kantorovich metric. This latter computation was done using the MCFZIB Minimum Cost Flow solver (Frangioni & Manca., 2006) for each pair of states and each action. The first of these methods will be referred to as *Kantorovich*. Cost reoptimization was used in the second of these exact metrics in order to speed up the computation, at the expense of larger space requirements. This second method will thus be referred to as *Kantorovich (reoptimization)*. The third method uses statistical sampling to approximate the transition distributions of each state. For all MDPs, 10 transition samples were taken for each state and action, and this vector of samples was used to estimate the empirical distribution throughout the whole run. The Hungarian algorithm was used to solve the assignment problem, as described in Section 3.3. The distance metric was obtained by averaging the distances obtained over 30 independent runs of this procedure. This method will be referred to as *Sample*.

The fourth method used the total variation metric; this provides a loose upper bound on the other metrics, but is much faster to compute. It will be referred to as *TV*. The fifth method uses the total variation metric with bisimilar equivalence classes, which provides a tighter upper bound. It will be referred to as *Bisim*. These metrics were computed using three different values for the discount factor: $\gamma = \{0.1, 0.5, 0.9\}$.

Table 1 summarizes the running times in seconds for each method with the different discount factors. A '-' means that the algorithm failed to compute the metric.

|  | Kantorovich | Kantorovich (reoptimization) | TV | Bisim | Sample |
|---|---|---|---|---|---|
| **3x3 gridWorld** | | | | | |
| $\gamma = 0.1$ | 2.067 | 1.563 | 0.028 | 0.743 | 5.574 |
| $\gamma = 0.5$ | 5.223 | 2.944 | 0.053 | 0.893 | 13.976 |
| $\gamma = 0.9$ | 41.089 | 15.231 | 0.303 | 2.533 | 84.624 |
| **5x5 gridWorld** | | | | | |
| $\gamma = 0.1$ | - | - | 0.341 | 103.906 | 42.395 |
| $\gamma = 0.5$ | - | - | 0.830 | 111.574 | 105.340 |
| $\gamma = 0.9$ | - | - | 5.715 | 190.018 | 627.370 |
| **7x7 gridWorld** | | | | | |
| $\gamma = 0.1$ | - | - | 2.181 | 2902.59 | 161.988 |
| $\gamma = 0.5$ | - | - | 5.768 | 3005.70 | 403.124 |
| $\gamma = 0.9$ | - | - | 42.023 | 4104.17 | 2408.97 |
| **Coffee Robot** | | | | | |
| $\gamma = 0.1$ | 57.640 | - | 0.254 | 14.637 | 57.708 |
| $\gamma = 0.5$ | 137.129 | - | 0.524 | 17.436 | 125.816 |
| $\gamma = 0.9$ | 1024.42 | - | 3.431 | 48.197 | 762.151 |

Table 1: Running times in seconds for different metrics

The amount of space used by each method was also compared. This was measured using the *massif* tool of valgrind (a tool library in Linux). Table 2 presents the maximum number of bytes used by each algorithm when computing the distances for the MDPs.

The distance metric computed is then used to reduce the state space of the original MDP by means of two aggrega-tion methods. In the first approach, the number of desired partitions is specified beforehand and a greedy, incremental aggregation procedure is used until the desired reduction has been reached. More precisely, each state starts as its own partition. Then, the algorithm greedily picks the two closest partitions and merges them. The distance between partitions is the minimum distance between pairs of states belonging to each partition. The value function is then computed for each state in the original MDP and each partition in the reduced MDP using value iteration. In Figures 1, 2, 3, and 4 we present the quality of the results obtained, measured as the $L_\infty$ norm of the value function error. More precisely, for each state, we compute the difference between its value in the original MDP, and its value as estimated in the aggregate MDP, and take the maximum absolute difference over all states, $\max_{s \in S} |V^*(s) - V^*([s])|$, where $[s]$ is the cluster containing $s$.

|  | Kantorovich | Reoptimize | TV | Bisim | Sample |
| --- | --- | --- | --- | --- | --- |
| 3x3 gridWorld | 80Mb | 180Mb | 80Kb | 70Kb | 80Kb |
| 5x5 gridWorld | 1.8Gb | 1.8Gb | 400Kb | 450Kb | 500Kb |
| 7x7 gridWorld | 1.8Gb | 1.8Gb | 1.2Mb | 1.6Mb | 1.8Mb |
| coffee robot | 1.6Gb | 1.8Gb | 300Kb | 325Kb | 300Kb |

Table 2: Memory usage in bytes for the different metrics

The second aggregation method greedily adds a state to a partition if its minimum distance to any state in the partition is less than ε. Both the process of creating partitions (by picking pairs of states to group) and adding states to partitions is greedy. The algorithm will stop when no more merging can be performed. Note that higher values of ε will lead to fewer partitions. The results are presented in Figures 5, 6, 7, and 8.

Based on the results presented here, the sampling method outperforms all the other methods when considering space, time and quality. The quality of the value functions obtained when using this metric for state aggregation is almost identical to the quality obtained for the "perfect" metric, $d^c_{fix}$. However, the memory requirements and time requirements are significantly smaller. The exact computation of the Kantorovich metric failed to finish for the two larger environments. Using cost reoptimization reduces the computation time by a factor of 2, on these small environments, but requires even more memory, so this method only completed for the smallest environment. The total variation metric, *TV*, is very cheap to compute but has very poor results in terms of the approximation error. The *Bisim* metric is slightly better than *TV*, but significantly worse than the sampling solution. However, the computation time and space are very similar to the ones of the sampling method.[7]

---

[7]Notice how the graphs plotting ε versus $L_\infty$ norm and size of aggregated MDP demonstrate that since the approximations are upper bounds, their $L_\infty$ norm appears 'better' than the exact methods, but at the expense of compression.

## 5 Conclusions and future work

In this paper, we discussed four state similarity metrics based on the notion of bisimulation. We compared and contrasted these both in theory and in practice. Based on these results, the metric obtained by means of sampling distributions, appears to be the clear winner: it significantly outperforms the other approaches when considering the tradeoff between the computational requirements of time and space, on one hand, and the quality of the results obtained when using this method for state aggregation, on the other hand. The next step is to test this metric on large-scale environments. Different versions of this metric, based on ideas from incremental reinforcement learning algorithms, rather than batch processing, will also be explored. Using such techniques, the computation could be made significantly faster. Versions of these metrics for factored state spaces are also of great interest.

The sampling approach is also promising for computing metrics in continuous state spaces. In prior works, we established the existence of of bisimulation metrics for continuous MDPs (Ferns et al., 2005). We are hopeful that the idea of approximating measures through empirical distributions will enable us to estimate the Kantorovich metric in a manner similar to the discrete case.

**Acknowledgments**

This work has been supported in part by funding from NSERC and CFI.

## Appendix A: Proofs

**Lemma 5.1.** *For any P and Q, $\{T_K^i(h)(P,Q)\}$ converges to $T_K(h)(P,Q)$ almost surely for any $h \in \mathcal{V}$.*

*Proof.* Let $\varepsilon > 0$. By the SLLN for each $x \in S$, $\{P_i(x)\}$ converges to $P(x)$ almost surely. In fact, since there are finitely many $x$ we have $\{P_i(x)\}$ converges to $P(x)$ almost surely for all $x$, and similarly for $Q$. So for each $x$ we may choose $i_x$ so large that

$$|P_i(x) - P(x)|, |Q_i(x) - Q(x)| < \frac{(1-c)\varepsilon}{|S|R} \text{ for all } i \geq i_x.$$

Define $i_0 = \max_{x \in S} i_x$. Then for all $i \geq i_0$

$$\begin{aligned}
&|T_K(h)(P,Q) - T_K^i(h)(P,Q)| \\
&= |T_K(h)(P,Q) - T_K(h)(P_i,Q_i)| \\
&\leq |T_K(h)(P,P_i) + T_K(h)(Q,Q_i)| \\
&\leq \frac{R}{1-c}(TV(P,P_i) + TV(Q,Q_i)) \\
&= \frac{R}{2(1-c)}(\sum_{x \in S}|P(x) - P_i(x)| + \sum_{x \in S}|Q(x) - Q_i(x)|) \\
&\leq \frac{R}{2(1-c)}(\sum_{x \in S}\frac{(1-c)\varepsilon}{|S|R} + \sum_{x \in S}\frac{(1-c)\varepsilon}{|S|R}) \\
&\leq \varepsilon
\end{aligned}$$

Here we have used the triangle inequality, monotonicity of the Kantorovich functional $T_K$, as well as the fact that $T_K$ applied to the discrete metric is $TV$. □

**Theorem 5.2.** *Given the setup of theorem 2.2, define $F_i^c : \mathcal{V} \to \mathcal{V}$ by*

$$F_i^c(h)(s,s') = \max_{a \in A}(|r_s^a - r_{s'}^a| + cT_K^i(h)(P_s^a, P_{s'}^a))$$

*Then:*

1. *$F_i^c$ has a unique fixed point $d_i^c$,*

2. *for any $h_0 \in \mathcal{V}$,*

$$\|d_i^c - (F_i^c)^n(h_0)\| \leq \frac{c^n}{1-c}\|F_i^c(h_0) - h_0\|,$$

3. *$d_i^c$ converges to $d_{fix}^c$ in uniform norm (almost surely).*

*Proof.* Note that the first two follow from theorem 2.2; by simply replacing the MDP probability parameters with the empirical distributions we obtain $d_i^c$ as the fixed point metric defined over the new MDP.

Let $\varepsilon > 0$. By lemma 5.1 we may choose $i$ so large that for any $x,y,a$, and $h$:

$$|T_K^i(h)(P_x^a, P_y^a) - T_K(h)(P_x^a, P_y^a)| < \frac{(1-c)\varepsilon}{c}.$$

Then

$$\begin{aligned}
&|d_i^c(x,y) - d_{fix}^c(x,y)| \\
&\leq c\max_a|T_K^i(d_i^c)(P_x^a, P_y^a) - T_K(d_{fix}^c)(P_x^a, P_y^a)| \\
&\leq c\max_a|T_K^i((d_i^c - d_{fix}^c) + d_{fix}^c)(P_x^a, P_y^a) - \\
&\qquad T_K(d_{fix}^c)(P_x^a, P_y^a)| \\
&\leq c(\|d_i^c - d_{fix}^c\| + \max_a|T_K^i(d_{fix}^c)(P_x^a, P_y^a) - \\
&\qquad T_K(d_{fix}^c)(P_x^a, P_y^a)|) \\
&\leq c(\|d_i^c - d_{fix}^c\| + \frac{(1-c)\varepsilon}{c})
\end{aligned}$$

Thus, we obtain $\|d_i^c - d_{fix}^c\| < \varepsilon$, as required. □

## Appendix B: Experiment Graphs

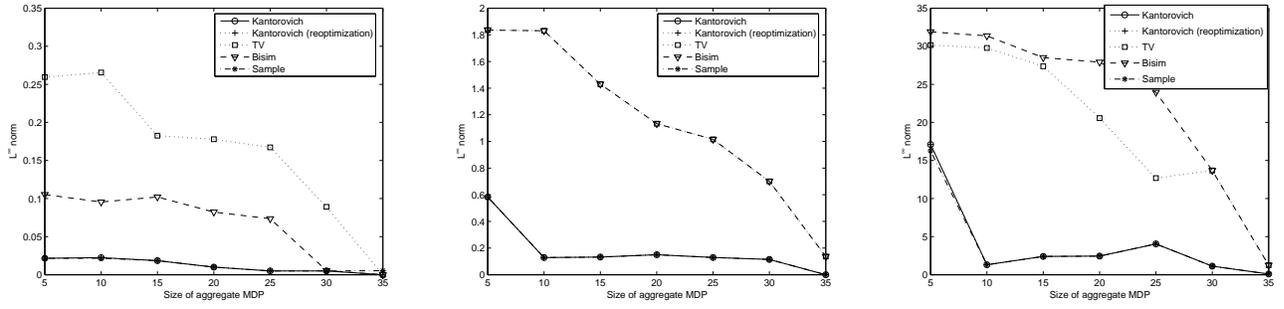

Figure 1: $L_\infty$ norm of 3x3 gridWorld with $\gamma = 0.1$ (left), $\gamma = 0.5$ (middle) and $\gamma = 0.9$ (right)

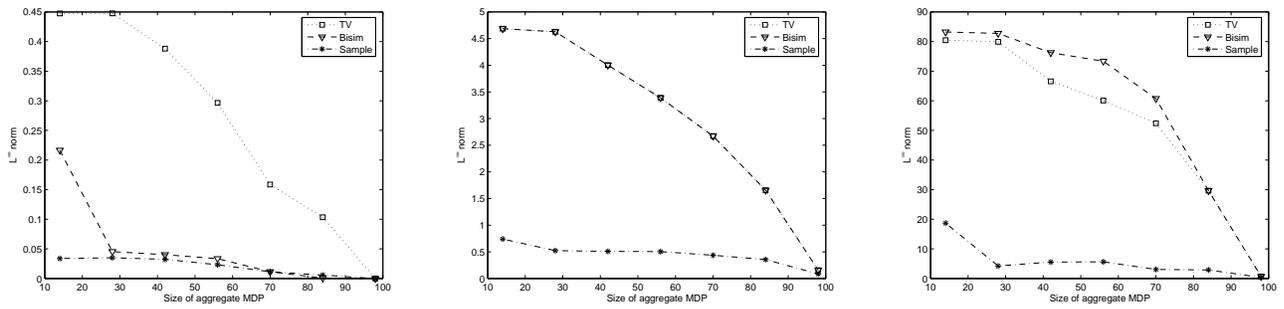

Figure 2: $L_\infty$ norm of 5x5 gridWorld with $\gamma = 0.1$ (left), $\gamma = 0.5$ (middle) and $\gamma = 0.9$ (right)

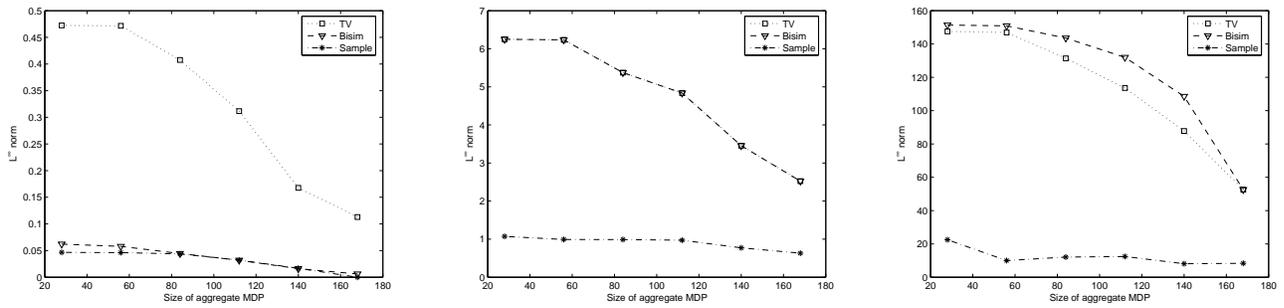

Figure 3: $L_\infty$ norm of 7x7 gridWorld with $\gamma = 0.1$ (left), $\gamma = 0.5$ (middle) and $\gamma = 0.9$ (right)

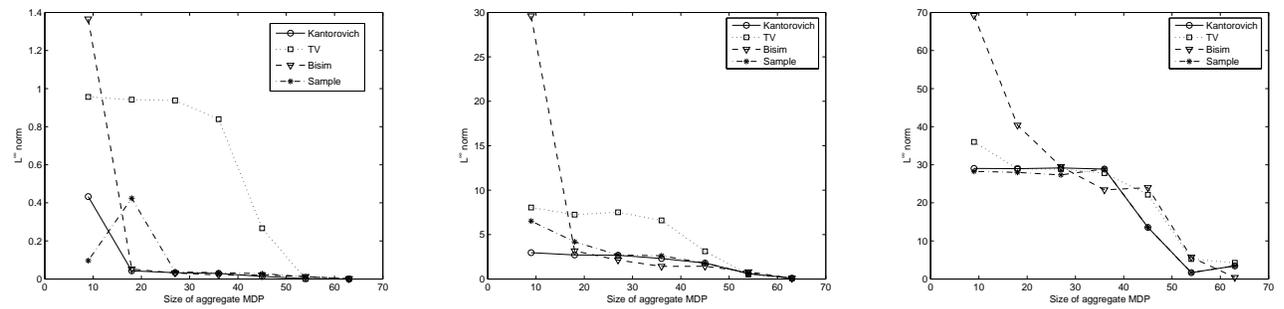

Figure 4: $L_\infty$ norm of coffee robot with $\gamma = 0.1$ (left), $\gamma = 0.5$ (middle) and $\gamma = 0.9$ (right)

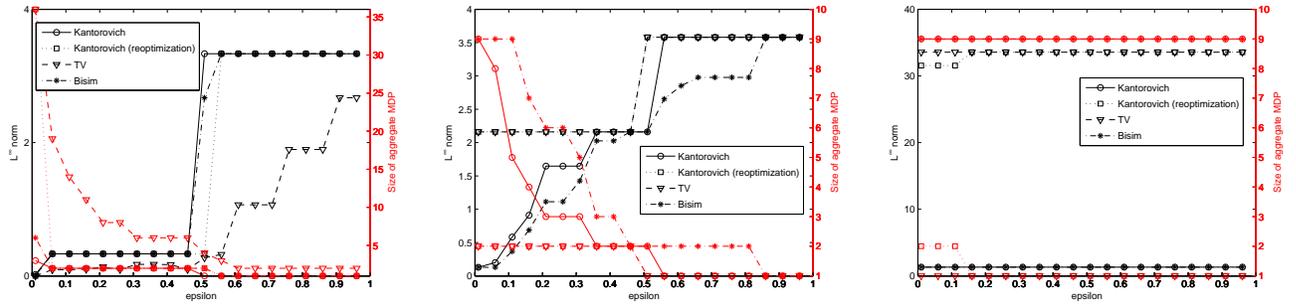

Figure 5: ε vs. $L_\infty$ norm (black) and size of aggregate MDP (red) of 3x3 gridWorld, with $\gamma = 0.1$ (left), $\gamma = 0.5$ (middle) and $\gamma = 0.9$ (right)

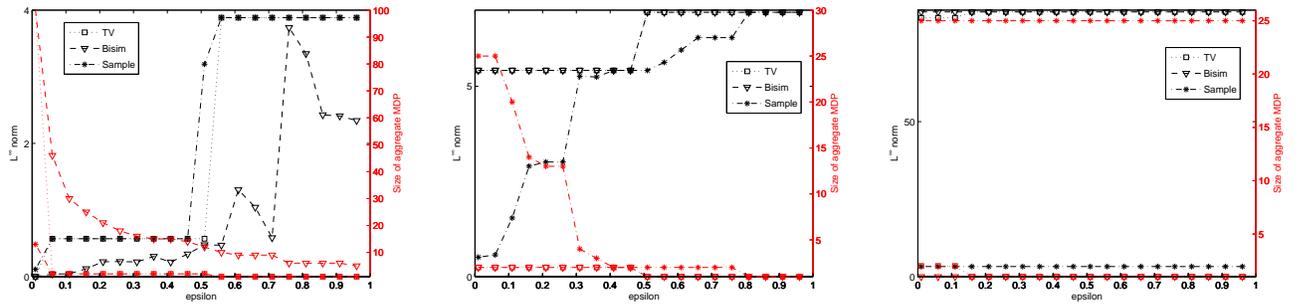

Figure 6: ε vs. $L_\infty$ norm (black) and size of aggregate MDP (red) of 5x5 gridWorld, with $\gamma = 0.1$ (left), $\gamma = 0.5$ (middle) and $\gamma = 0.9$ (right)

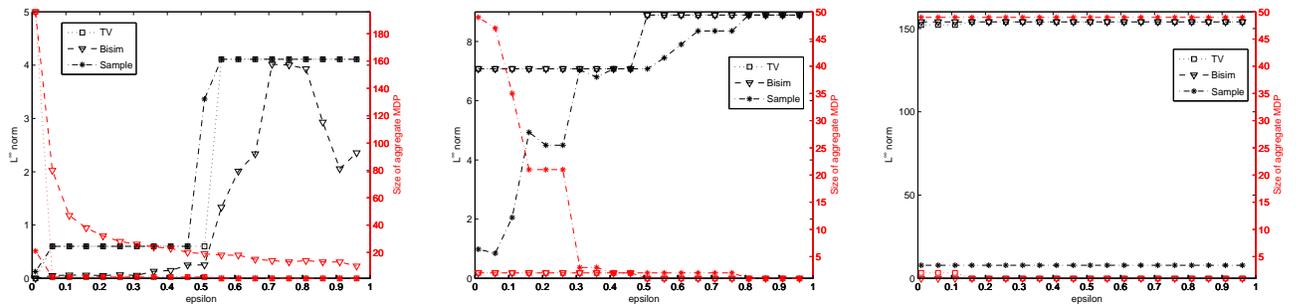

Figure 7: ε vs. $L_\infty$ norm (black) and size of aggregate MDP (red) of 7x7 gridWorld, with $\gamma = 0.1$ (left), $\gamma = 0.5$ (middle) and $\gamma = 0.9$ (right)

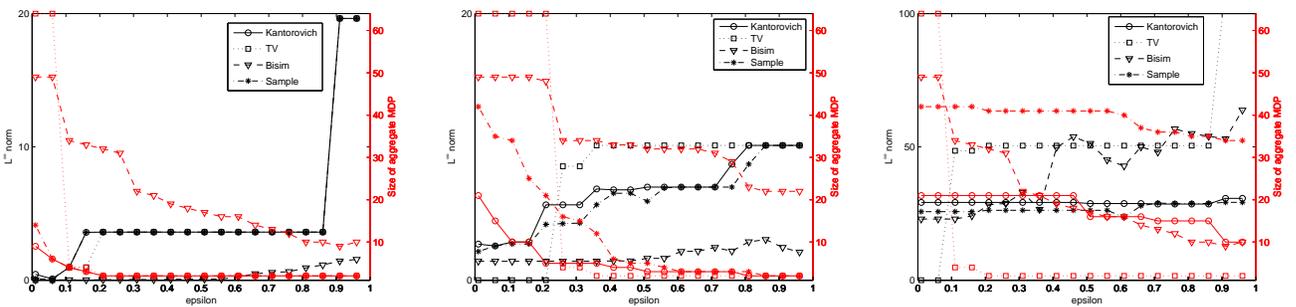

Figure 8: ε vs. $L_\infty$ norm (black) and size of aggregate MDP (red) of coffee robot, with $\gamma = 0.1$ (left), $\gamma = 0.5$ (middle) and $\gamma = 0.9$ (right)